\documentclass[runningheads,a4paper]{llncs}

\usepackage{amssymb}
\usepackage{amsmath}
\usepackage{graphicx}
\usepackage{wrapfig}

\usepackage[sort, numbers]{natbib}
\usepackage{todonotes}

\usepackage{url}
\urldef{\mailsa}\path|{alfred.hofmann, ursula.barth, ingrid.haas, frank.holzwarth,|
\urldef{\mailsb}\path|anna.kramer, leonie.kunz, christine.reiss, nicole.sator,|
\urldef{\mailsc}\path|erika.siebert-cole, peter.strasser, lncs}@springer.com|

\begin{document}

\mainmatter  

\title{Integrating Deep Learning in Domain Sciences at Exascale
\footnote{
This manuscript has been authored by UT-Battelle, LLC, under contract DE-AC05-00OR22725 with the US Department of Energy (DOE). The US government retains and the publisher, by accepting the article for publication, acknowledges that the US government retains a nonexclusive, paid-up, irrevocable, worldwide license to publish or reproduce the published form of this manuscript, or allow others to do so, for US government purposes. DOE will provide public access to these results of federally sponsored research in accordance with the DOE Public Access Plan (http://energy.gov/downloads/doe-public-access-plan).
}}

\titlerunning{Integrating Deep Learning in Domain Sciences at Exascale}


\author{
    Rick Archibald \inst{1} \and
    Edmond Chow \inst{2} \and
    Eduardo D'Azevedo \inst{1} \and
    Jack Dongarra \inst{3,1} \and
    Markus Eisenbach \inst{1} \and
    Rocco Febbo \inst{3} \and
    Florent Lopez \inst{3}\and
    Daniel Nichols \inst{3} \and \\
    Stanimire Tomov \inst{3} \and
    Kwai Wong \inst{3} \and
    Junqi Yin \inst{1}
}

\authorrunning{Archibald et. al.}


\institute{
    Oak Ridge National Laboratory \\
        \email{\{archbaldrk, dazevedoef, eisenbackm, yinj\}@ornl.gov}  \and
    Georgia Institute of Technology \\
        \email{echow@cc.gatech.edu} \and
    University of Tennessee, Knoxville \\
        \email{\{dongarra, flopez, tomov\}@icl.utk.edu} \\
        \email{\{rfebbo, dnicho22\}@vols.utk.edu} \\
        \email{kwong@utk.edu}
}

\toctitle{TOC Title}
\tocauthor{TOC Author}
\maketitle

\begin{abstract}
This paper presents some of the current challenges in designing deep learning artificial intelligence (AI) and integrating it with traditional high-performance computing (HPC) simulations. We evaluate existing packages for their ability to run deep learning models and applications on large-scale HPC systems efficiently, identify challenges, and propose new asynchronous parallelization and optimization techniques for current large-scale heterogeneous systems and upcoming exascale systems.
These developments, along with existing HPC AI software capabilities, have been integrated into MagmaDNN, an open-source HPC deep learning framework.
Many deep learning frameworks are targeted at data scientists and fall short in providing quality integration into existing HPC workflows. This paper discusses the necessities of an HPC deep learning framework and how those needs can be provided (e.g., as in MagmaDNN) through a deep integration with existing HPC libraries, such as MAGMA and its modular memory management, MPI, CuBLAS, CuDNN, MKL, and HIP. Advancements are also illustrated through the use of algorithmic enhancements in reduced- and mixed-precision, as well as asynchronous optimization methods.
Finally, we present illustrations and potential solutions for enhancing traditional compute- and data-intensive applications at ORNL and UTK with AI. The approaches and future challenges are illustrated in materials science, imaging, and climate applications.


\end{abstract}

\section{Background}

\vspace{-0.1in}
Deep learning (DL) has become one of the most prevalent technologies today. Applications extend from image recognition to natural language processing. While its commercial applications have garnered the efforts of major technology companies, DL has proven vital to works in numerous scientific domains. 

Some of these domain problems are vital to solving many of today's global issues---for instance, in solving modern, data-dense problems in climate science. This field has seen several advancements with the recent integration of deep neural networks (DNNs). Ise et. al. \cite{10.3389/frobt.2019.00032} propose a modern approach towards utilizing neural networks to predict global temperature data. They further conclude that the accuracy of their model increases with the number of input images used. Thus, the model's accuracy and utility grow with the amount of compute power available. Wang et. al. \cite{gmd-12-4261-2019} provide further uses for DNNs in weather forecasting, which compete with the current state-of-the-art, physically based simulations. The neural networks can also reduce the computations necessary, as they possess the ability to extend one trained network to similar environments, or train once, and predict frequently. 

Additionally, DL has proven useful in materials science, even leading the field of materials informatics \cite{agrawal_choudhary_2019}. Feng et. al. \cite{Feng_Zhou_Dong_2019} were able to utilize deep networks to predict material defects with an accuracy above 98\%, even for small data sets. However, these great accuracies require deep and/or pre-trained networks. As with climate modelling, deep learning can be used to replace current physically based simulations, which are expensive and/or do not scale well~\cite{Ye_Chen_Wang_Chu_Ong_2018}.

Providing high performance hardware and software stacks to facilitate deep learning has become a focus in the high-performance computing (HPC) field.  An overview of the parallel algorithms and their challenges for distributed, deep neural networks can be found in Ben-Nun and Toefler \cite{Demystifying-DL}. Large-scale modern systems, such as the US DOE's Summit, are fully loaded with high-performance GPUs to support the compute-intensive nature of deep learning. Additionally, some HPC distributed file systems under-perform in providing I/O for data-hungry DL training \cite{8890993}. Even in terms of sheer performance, many state-of-the-art deep learning models and libraries fail to utilize a significant portion of the maximum floating-point operations (FLOPs) rate of current top systems \cite{10.1145/3225058.3225069}. There is significant research in the area of accelerating DL on cloud and other commercial systems, but these methods may not integrate well with varying HPC technologies and modern scientific workflows. 

One major area in which HPC can innovate deep learning is model parallelism. In many domain applications, networks can become large, leading to growth beyond the memory of a single GPU. By utilizing existing HPC communication standards, a robust model-parallel training standard needs to be developed in addition to efficient implementations. Currently, several libraries attempt to integrate model parallelism \cite{mesh-tf}, but this requires extensive knowledge of the code base, model, and hardware. There is an active research push into further developing model parallel approaches \cite{10.1145/3322795.3331463, Huang_Cheng_Bapna_Firat_Chen_Chen_Lee_Ngiam_Le_Wu, 10.1145/3210377.3210394, Chen2018EfficientAR}.

R. Stevens presented a vision of the landscape of machine learning as DOE moves towards the exascale era~\cite{RStevens:18}, stressing the importance of machine learning as an integral part of DOE exascale applications. The new paradigm of "HPC + AI" on exascale computing leads towards an "Integrated Sim, Data, Learn Stack." There has been a long history of application code development for compute-intensive applications in the areas of climate, materials, medical, high energy, and urban sciences applications. In light of these needs, we will highlight developments and discuss the necessity of a set of tools and methods particularly relevant to deploying AI on HPC systems.

Realizing the paradigm of ``HPC + AI" requires three major components: (1)~the HPC system and investments in scaling application-driven capabilities; (2) the AI software achieving the potential performance; and (3) the ``+", innovative algorithms and implementations/tools that combine the components of compute- and data-driven applications seamlessly on exascale machines. 

\section{Deep Learning Software on Modern HPC Systems}

There are countless deep learning frameworks in existence today: TensorFlow~\cite{tensorflow}, PyTorch \cite{Paszke_Gross_Massa_Lerer_Bradbury_Chanan_Killeen_Lin_Gimelshein_Antiga}, MxNet \cite{mxnet:15}, and many others. Each framework has its own advantages and each falls short in some areas. One advantage of each of these frameworks is their strong corporate backings. Companies pour significant amounts of money into making production-ready libraries, which support various hardware. However, when using these frameworks in an HPC setting, this can cause issues. Much of the research and advancements by the library designers are production focused and target cloud environments. Some of these advancements translate well onto HPC systems, such as the use of hardware accelerators and mixed precision. However, many communication paradigms and implementations differ dramatically between the two. For example, the parameter server method in TensorFlow is more suited for heterogeneous cloud environments, although it is more bandwidth efficient than allreduce; and the socket-based GLOO communication backend built into PyTorch cannot efficiently utilize high-performance interconnects on HPC systems, although it is more flexible.      

\subsection{Towards a Deep Learning Framework for HPC}
\begin{wrapfigure}{r}{0.45\linewidth}
  \centering
  \vspace{-0.1in}
  \includegraphics[width=1\linewidth]{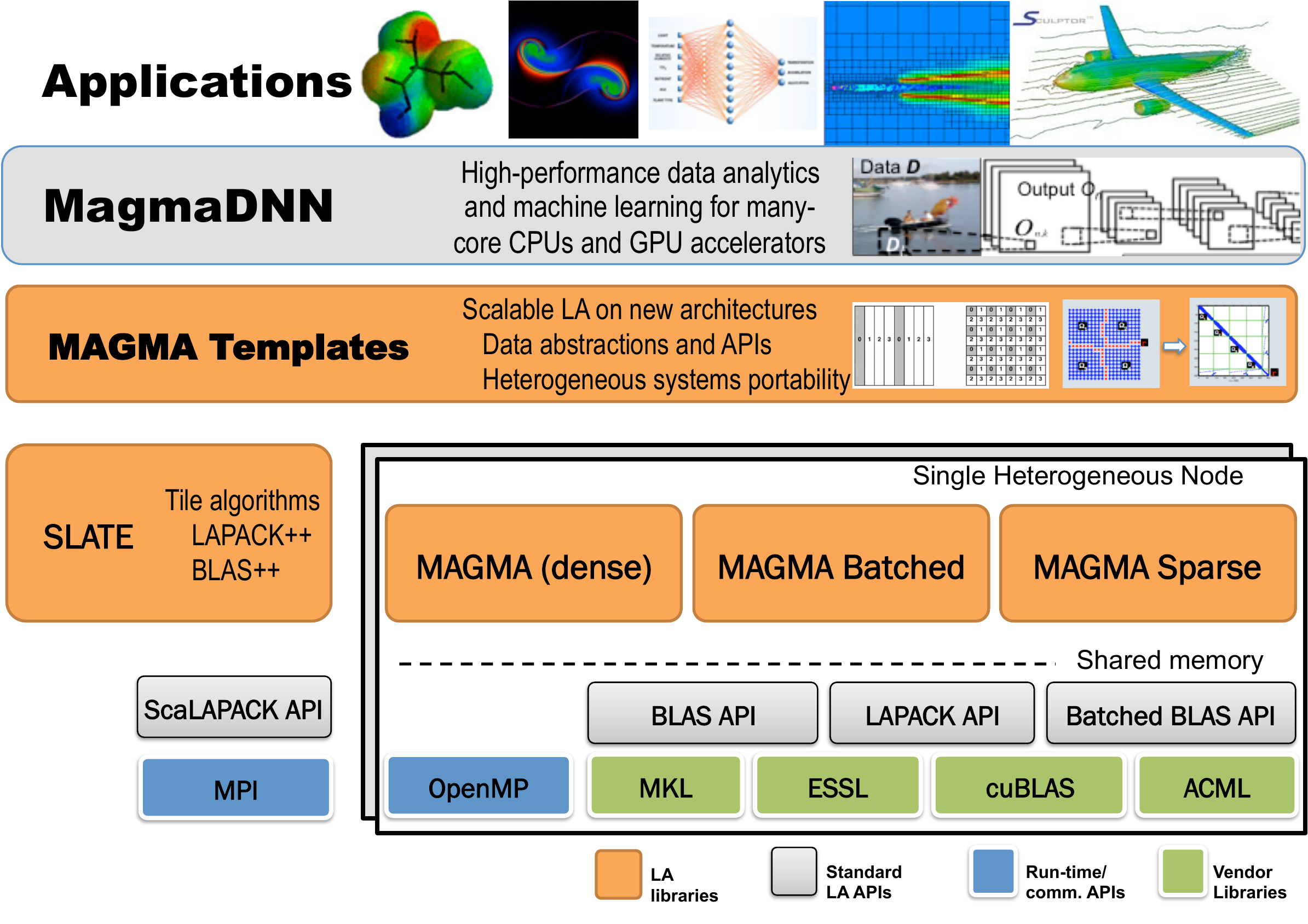}
  \caption{MagmaDNN software stack.}
  \vspace{-0.2in}
  \label{fig:magmadnn}
\end{wrapfigure}

We evaluated a number of existing packages for their readiness to efficiently run deep learning models and applications on large-scale HPC systems. As	pointed	out previously, we	found various limitations. Our vision of what is important and needed in deep leaning AI packages for HPC is incorporated into the design of the MagmaDNN open source framework~\cite{10.1145/3332186.3333047,magmadnnISC19} (see Figure~\ref{fig:magmadnn}) with current release MagmaDNN v1.2~\cite{daniel_nichols_2020_3972406}, which we discuss briefly in what follows.

Firstly, MagmaDNN targets HPC systems, rather than cloud systems. It is written in C++ and is open source (available at \url{bitbucket.org/icl/magmadnn}).

Many of the core components of MagmaDNN are the same as other frameworks: operations are represented in a compute graph, data is passed around via tensors, gradients are computed using backpropagation, etc. However, each of these was designed with an HPC environment in mind.

MagmaDNN is centered around existing high-performance software packages, enabling easy integration into existing environments and Exascale Computing Project (ECP) software products. First and foremost, MagmaDNN uses the Matrix Algebra on GPU and Multicore Architectures (MAGMA) \cite{magma} package for heterogeneous linear algebra, since deep learning is heavily dependent on linear algebra routines. For strictly CPU computations, the package allows the use of any Basic Linear Algebra Subprograms (BLAS)/LAPACK packages such as Intel MKL, OpenBLAS, ATLAS, etc. 

To better operate with hardware accelerators, MagmaDNN uses CUDA packages to harness GPU acceleration. Custom CUDA kernels, along with CUDNN's routines for GPU-accelerated deep learning, allows the package to get maximal performance on GPU-enabled systems.

For distributed computing, MagmaDNN relies on existing HPC tools such as the Message Passing Interface (MPI) for its communication. Most systems have an MPI distribution tuned to that hardware, while some even have custom distributions. Relying on MPI for inter-node communication allows the framework to utilize existing optimized implementations while providing simpler integration with existing scientific codes.

This integration of existing HPC frameworks allows MagmaDNN to compile and link with the fastest libraries for particular hardware. Utilizing existing optimized packages is a consistent design pattern within HPC libraries, and MagmaDNN uses interfaces that are commonly installed on HPC clusters. In terms of integrating with existing systems and scientific codes, this approach offers the package a significant advantage over other frameworks and workflows.

Not only does MagmaDNN integrate with current HPC frameworks, it provides modularity in its design and linking so that users have flexibility in use. Conforming to standard APIs allows linking with any choice of framework; any BLAS/LAPACK/MPI choice implementations can be used. However, it is not just libraries with which MagmaDNN is modular. By utilizing computation graphs and extendable C++ classes, the package allows users to fully customize their neural networks. For instance, to utilize a unique network interface for distributed training, one can define a custom node on the computation graph which handles the transfer of data on that network component. This can be further utilized to define any desired data movement or computational elements. By providing modular components, MagmaDNN provides a structure onto which domain scientists can easily implement desired custom functionality.

Much of this customizability is fully supported through MagmaDNN's interfaces. For instance, working with GPU and distributed memory is abstracted away in the \verb|MemoryManager| class. Thus, library users do not need to worry about where memory is stored and how it is transferred. The package identifies the memory type, and whenever a memory operation is used (e.g., copy, put, get, etc.), and the MagmaDNN implementation handles the low-level specifics. Likewise, communication operations between devices or nodes are abstracted into the library as well. However, users can define their own operations, meaning they can redefine those in existing code. This makes it easy, for example, to try out different convolution algorithms and determine which one performs optimally. Another common example is to determine which distribution strategy is optimal for a specific machine. The same training code will run on any machine, and only the communication code might need to be altered.

Looking towards next-generation hardware and exascale systems, MagmaDNN aims to incorporate more technology stacks such as AMD's HIP (e.g., hipMAGMA is already available) and Intel's OneAPI. Support for more technologies will allow researchers at various labs and institutions to focus on domain research---and not on how to integrate their code with HPC stacks. We also hope to provide streamlined and optimal support for model parallelism. As the size of models grows, so does the necessary memory requirements on devices. Soon many cutting-edge networks, which can provide state-of-the-art accuracy in classification, will require model parallelism. Supporting fast, scalable model parallelism on HPC systems is crucial to facilitating next-generation research, and MagmaDNN is moving towards this goal.

\subsection{Workflow Software for Modern HPC Systems}

\begin{figure}[ht]
    \hspace{0.2in}
    \begin{minipage}[b]{0.35\linewidth}
    \centering
    \includegraphics[height=1.5in, width=1.5in]{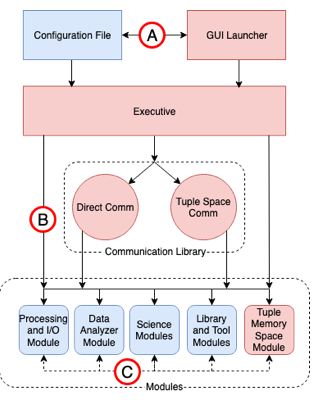}
    \caption{OpenDIEL Structure.}
    \label{fig:minipage1}
    \end{minipage}
    \quad
    \begin{minipage}[b]{0.45\linewidth}
    \centering
    \includegraphics[height=1.5in, width=2.6in]{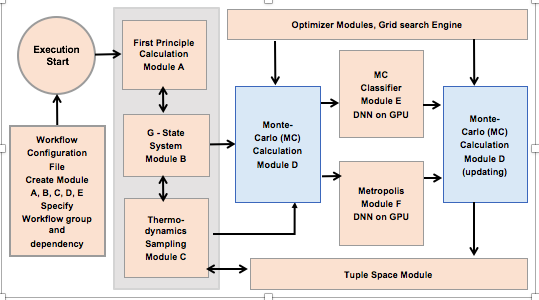}
    \caption{Workflow in Materials Sciences.}
    \label{fig:minipage2}
    \end{minipage}
    \end{figure}

Another major component in the integration of AI and HPC is the availability of workflow software: which is critical, for example, in launching, collecting results from, and analyzing various simulations in hyperparameter tuning, DL networks discoveries, etc.
Exascale applications largely have their own set of workflow procedures and tools to launch their simulations. Integrating traditional applications seamlessly with existing AI software---either TensorFlow, PyTorch, or MagmaDNN---takes on a new level of challenges. Most workflow frameworks available today focus on cloud deployment, yet a workflow framework tailored to scale on HPC systems is critical to the success to "Automate and Accelerate" the task of "Sim, Data, and Learn Stack". SWIFT-T \cite{swiftt} is a tool that offers the choice to run on large-scale HPC systems. At The University of Tennessee, Knoxville (UTK), we have developed a parallel workflow framework called
openDIEL that aims to give researchers
and users of HPC machines an efficient way to coordinate, organize,
and interconnect many disparate modules of computation in order to effectively utilize and allocate HPC resources \cite{opendiel18,opendiel19}.
It provides users an adaptive avenue to run
compute-intensive and data science jobs concurrently, allowing specification of DNN architectures, data processing, and hyperparameter
tuning. Existing ML tools can be readily used
in openDIEL, allowing for easy experimentation with
various models and approaches.  Most importantly, openDIEL provides a platform to run existing ECP applications in their own workflow procedures.
When conducting multi-discipline simulations, it is often required
to use a large variety of software, and serial or parallel codes, to answer
a research question. Utilizing disparate modules of computation
requires the careful coordination of dependencies,
communication, and synchronization between them, and
there is not always a clear path. 
This problem is solved by openDIEL: it enables
researchers to define complex dependencies between modules
and schedule communication between them. 
OpenDIEL consists of three primary components: the workflow control
file, the communication library (COMMLIB), and the
Executive, shown in Figure~\ref{fig:minipage1}. 
Workflow is defined by users via the control file, which consists of two parts, the definition of functional modules and the interactions and sequence of execution of modules. OpenDIEL provides two communication libraries, one for direct point-to-point data transfer between two modules, and a store-and-move tuple space library for collective and asynchronous data transfers. The Executive is a lightweight manager designed to run and scale on heterogeneous HPC platforms. 

\vspace{-0.15in}
\vspace{-0.1in}

OpenDIEL
specifically focuses on unifying modules into a single executable
that uses MPI for communication. 
Climate and materials sciences simulations, with typical workflow shown in Figure~\ref{fig:minipage2}, are two of the applications that are incorporating deep learning into their simulations.

\section{Algorithmic Improvements for DNN AI in HPC}
In this section, we select and highlight the development of two technologies that we believe are critical for the scalability and efficient use of HPC systems. The first (Section~\ref{sec:asgd}) addresses scalability of DL training, stressing the benefits and need for asynchronous methods on today's highly parallel and heterogeneous systems, featuring
powerful nodes and relatively slow interconnects. The second (Section~\ref{sec:mixed}) addresses hardware changes motivated by DL---in particular powerful hardware acceleration for low-precision arithmetic---and techniques for harnessing this in scientific HPC applications.

\subsection{Asynchronous Methods}\label{sec:asgd}

In a multi-core or many-core shared-memory environment, the speed at which
the parameters can be updated by different threads may be a limiting 
factor for stochastic gradient descent (SGD) performance.  This is because threads must lock the 
parameter vector while they are updating it, preventing other threads from
reading or updating the parameters, so that they must wait.  Asynchronous 
versions of SGD may relax the data dependencies and order of operations
compared to the classical SGD algorithm, allowing the algorithm to update parameters 
more rapidly---but also making the asynchronous algorithm mathematically different
and non-deterministic in its execution \cite{hogwild:11}.

In a distributed-memory environment, a common implementation option for
SGD is for each compute node to store a copy of the model parameters
\cite{horovod}.  When each compute node has computed a gradient for its
mini batch of data, all the gradients must be summed and then shared
among the compute nodes so that they can update their copies of the
model parameters.  This leads to a potential bottleneck in the global
all-reduce computation at every step of SGD.  In addition, nodes that
compute their gradients faster than others must wait for the other
nodes (sometimes called ``stragglers'').  This can be a very
high synchronization overhead that may be difficult to reduce by
load balancing on heterogeneous architectures.  An asynchronous
approach can address these high overheads, as we demonstrate below.

Another common implementation option for distributed-memory computers
is the parameter server approach \cite{distbelief:12}.  
Here, one or more compute nodes are dedicated
to storing and updating the model parameters.  The worker nodes compute
gradients for their partition of the data and send these gradients
to the parameter server.  The parameter server then sends back the updated
values of the model parameters.  The parameter server model can be
used to run SGD synchronously---with the same issues of stragglers
slowing down the iteration for all nodes---but it has the advantage that it
can also be run asynchronously (i.e., the worker nodes do not have
to send gradients at the same time).  Workers, however, may be working
with different sets of values for the parameters at any instant.
This makes the computation asynchronous.
In distributed memory implementations, the gradients are 
typically updated by the parameter
server using atomic memory operations, which differs from the shared memory case.


Asynchronous parameter server implementations were proposed around 2012 or
earlier, but it was Hogwild! \cite{hogwild:11}, with inconsistent reads and writes of
the parameter vector in shared memory, that popularized the asynchronous
approach.  In particular, Hogwild! showed that, at least in cases of convex
problems and gradient sparsity, asynchrony with inconsistent reads and writes
does not have to harm convergence asymptotically.  Hogwild! was extended
in many directions, such as
HogBatch \cite{hogbatch}, designed for efficiently running on multi-core CPU architectures,
and Buckwild~\cite{s.z.o.r:15} which exploits low-precision arithmetic
to accelerate the training. Improvements on the analysis of Hogwild! showed
that convergence can be maintained for the non-convex
and non-sparse case~\cite{NIPS2015_5751}.
In MagmaDNN, we designed a parallel asynchronous variant of SGD that is similar
to a parameter server approach, but on multi-core
CPU and GPU architectures~\cite{l.c.t.d:20}.

In practice, the accuracy of machine learning models can be dramatically
improved by increasing the number of model parameters as well as the
sizes of the training sets. As a result, the memory footprint needed
for model training often exceeds the memory storage available on a
single node. In such cases, models must be trained on large-scale
machines where the dataset, and possibly the model as well, are
distributed among the compute nodes. From a performance perspective,
it is crucial to efficiently parallelize the training in this setup---and this means overcoming the communication costs. It has been shown
that when targeting large-scale platforms, asynchronous algorithms can
outperform the synchronous ones as in the shared-memory case. In
Google's DistBelief framework~\cite{distbelief:12}, for example, the
Downpour SGD algorithm uses a centralized parameter server for storing
the model, and distributes the dataset to other participating
nodes. Asynchrony enables communications to overlap with computations,
thus improving resource usage. In addition, the algorithm is
particularly well adapted to heterogeneous computing environments
where resources may have different processing and communication
speeds.

In MagmaDNN, we take a different approach,
exploiting remote memory access (RMA) using MPI one-sided
communication capabilities for implementing asynchronous SGD in order
to maximize the scalability of parallel training. This approach has
already been proven efficient in implementing several asynchronous
numerical algorithms such as the Jacobi~\cite{jacobi-async:19} or
optimized Schwarz~\cite{optimized-schartz:19} methods for solving
sparse systems of linear equations.

\subsection{Reduced and Mixed Precision}\label{sec:mixed}

Deep neural networks can be efficiently trained using low-precision floating-point arithmetic, such as single (fp32) or lower precision, improving the training times on modern hardware~\cite{lowprec-training:14}. Popular floating-point formats include the 16-bit IEEE half-precision format (fp16) and the 16-bit bfloat16~\cite{inte18} format. With fewer bits in the mantissa compared to fp16, the bfloat16 format offers less precision but has the advantage of having the same range as fp32---thereby removing the risk of overflow and underflow when converting from fp32 data.

Motivated by the success of low-precision arithmetic in machine learning applications, many specialized hardware accelerators have been designed with reduced- or mixed-precision arithmetic capabilities. For example, the NVIDIA Tensor Cores introduced in the Volta GPU can issue $4 \times 4$ 
matrix multiplication $D = C + A * B$ in one clock cycle, where the input matrices
$A$ and $B$ are fp16, whereas $D$ and $C$ can be either fp16 or fp32. The theoretical performance peak on Volta GPU is 125 teraFLOP/s, which is 8 times the performance peak
for fp32; and, although the accuracy of tensor core operations is lower than fp32, the ability to accumulate the result of matrix multiplication in fp32 rather than fp16 yields a considerable gain with respect to accuracy over fp16~\cite{blockfma:20}. In~\cite{Azzam_FP16}, authors use Tensor Cores for computing a reduced-precision LU factorization of a dense linear system, initially generated in double precision (fp64), and use these factors as a preconditioner for a Krylov solver in order to retrieve an fp64 solution. By doing this, they manage to reduce the time to solution up to $4 \times$ fp64 arithmetic for the factorization.  

Tensor Cores can also be used in performing FFTs~\cite{Kwai_FFT}.
Note that a matrix in fp32 can be well approximated as the scaled sum of two fp16 matrices
\begin{equation}
A_{32} \approx a1_{32} * A1_{16} + a2_{32} * A2_{16}
\label{eqn:fp32_fp16}
\end{equation}
where $a1_{32}$ and $a2_{32}$ are in fp32.
Scaling by $a1_{32} (a2_{32})$ ensures $A1_{16}  (A2_{16})$ is within the
limited dynamic range of fp16. 
Conceptually, $A1_{16}$ roughly captures the 3 most significant decimal digits, and $A2_{16}$ the
next 3 lower significant decimal digits.
If $A_{32}$ is already well scaled, then one can expect $a1_{32} = 1$ and $a2_{32}=2^{-11} * a1_{32}$.
Matrix multiplication of two fp32 matrices can be approximated by 3 matrix multiplications of
fp16 matrices on Tensor Cores, which may give a theoretical speedup of 8/3$\times$ on Volta GPU.
For another perspective, equation (\ref{eqn:fp32_fp16}) suggests the form of operations in DNN are theoretically able to well approximate operations of fp32 using mixed-precision operations in fp16.

\section{Applications}

\subsection{Materials Science and Microscopy}
There are multiple opportunities to exploit machine learning techniques in materials science. 
While many applications have concentrated on materials discovery---as exemplified by the use of ML in conjunction with databases such as The Materials Project \cite{Persson2013}, AFLOW \cite{Curtarolo2018}, or OQMD \cite{Wolverton2015}---a tightly coupled integration of traditional simulation techniques has great promise for  bridging the accuracy and computational cost divide between first principles calculations and effective models.
One promising combination for a tight coupling is for first principles statistical mechanics of materials to calculate the temperature dependence of materials. This requires the calculations of many possible atomic configurations within these materials using a Monte Carlo approach, where the probability of the individual states would be evaluated using an expensive density functional theory calculation~\cite{Eisenbach2009}. Thus, we will utilize ML to construct surrogate models as an intermediate step to link the first principles calculations to the Monte Carlo simulation.
Here, the calculation of ordering phase transitions in alloys might serve as an example \cite{Eisenbach2019}. In a solid solution alloy, different chemical species can occupy the sites of an underlying lattice structure (note that the total number of states grows exponentially with the number of atoms in the system). Each of these possible configurations has a different energy that determines the probability of the system being found in this state at a given temperature. To build a model for these interactions, density functional calculations for representative configurations O(1,000--10,000) will be performed to train a surrogate model that can replace the expensive (in the order of multiple node hours per data point) first principles calculation within the Monte Carlo sampling of possible configurations. While this approach can be conducted in a linear workflow, $\textrm{DFT}\rightarrow\textrm{ML}\rightarrow\textrm{MC}$, we envision a tighter coupled workflow, which augments the original training set with new points from important regions of the phase space discovered during the MC simulation, and retrains the model to improve the quantitative predictive power of this approach.

A long-standing inverse problem in atomic imaging is the loss of phase information during measurement (a.k.a., the phase problem). Given the sparse data collection of scanning transmission electron microscopic (STEM) images on different types of materials, a comprehensive database is needed for the community to study the subject. State-of-the-art electron microscopes produce focused electron beams with atomic dimensions and capture of convergent beam electron diffraction (CBED) patterns. In this dataset \cite{stemdldata}, we use newly developed electron scattering simulation codes to generate CBED patterns from over 60,000 materials (solid-state materials) from a material project database, representing nearly every known crystal structure. A data sample from this data set is given by a 3D array formed by stacking 3 CBED patterns simulated from the same material at 3 distinct material projections (i.e., crystallographic orientations). Associated with each data sample in the data set is a host of material attributes or properties which are, in principle, retrievable via analysis of this CBED stack. These consists of the crystal space group to which the material belongs, atomic lattice constants and angles, and chemical composition, to name but a few. Of note is the crystal space group attribute (or label). 

This dataset could be augmented with experimental data in the future. The generated dataset, based on simulations, will emphasize the scalability aspect of the model with the use of very large images ($\sim$ 10GB per image).   


\subsection{Super-Resolution for HPC Simulations}
Image compression is a very active field of study, with new methods being constantly generated \cite{sayood2017introduction}.  The need for improvements in image compression quality is growing in the field of HPC simulations because of the exponential trend in data generation.  There exists an untapped potential in this situation due to the nature of simulated data that is not currently exploited.  Simulation data from numerical systems of partial differential equations exist on a solution manifold~\cite{sol_man}.  Thus, the manifold hypothesis in machine learning---which states that real-world, high-dimensional data lie on low-dimensional manifolds embedded within the high-dimensional space---is concrete for simulation data.  We can therefore expect that identifying this map to the low-dimensional manifold will provide ideal compression for HPC Simulations. In the next paragraph we describe the basic setup that allows researchers to test in situ machine learning for climate simulations.    
\par
The shallow water equations on a sphere is a well-known model used in climate science to simulate basic dynamics of the Earth's atmosphere.  There exist many test case scenarios for this model \cite{SW_Test,NJ08,OldMcDonald,GSP04}, and in this paper we use the challenging test known as the barotropic instability test \cite{GSP04}, which involves perturbing the atmospheric flow with a localized bump to the balanced height field. We simulate the shallow water equations on the sphere by using a fourth-order Runge-Kutta method \cite{RK}  for time-stepping and a discontinuous Galerkin (DG) spatial discretization with a Legendre basis on a cubed-sphere mesh \cite{CubeSphere,DG_CS_Trans2}.
\par
The data structure for this simulation consists of a time series of images for each of the six faces of the cube, which for analysis can be converted to a single series lat-lon gridded images. Arguably, the most commonly used form of lossy compression is the discrete cosine transform (DCT) method \cite{1672377}. 
ECP has two projects on the development of FFTs and related transformations, one of which is heFFTe~\cite{heffte0.2}, already integrated in the ECP software stack, delivering high-performance (close to 90\% of the theoretical roofline peak), and very good weak and strong scaling on systems like Summit at ORNL. 
This common compression method fills out the setup of the in situ machine learning super resolution method.  At each time step of the barotropic instability simulation, in situ machine learning methods are exposed to both the compressed image data and the original image.  As the simulation progresses, only the compressed image data is stored to disk, and the machine learning method adaptively learns correct super-resolution transformation of the compressed data. Final stored data contain all compressed image data with the trained machine learning method.  It has been demonstrated in \cite{9006550} that, using this setup, it is possible to train in situ networks to reduce the error in lossy compression---obtaining multiple orders of improvement.  Going forward, it will be necessary to further improve in situ compression and analysis in order to maximize discovery with HPC simulations.  

\section{Meeting Exascale}

Exascale systems will appear in 2021. These machines incorporate accelerators from NVIDIA, AMD, and Intel. Efforts in preparing the software stack for exascale systems have been a major focus in the DOE ECP program~\cite{ECPSite}. We believe a native DNN framework such as MagmaDNN will pave a unique path to meet the challenges of exascale platforms. Assisted by the openDIEL parallel workflow engine, which admits a diverse spectrum of applications as functional modules, we will be able to exploit the full capabilities of exascale machines.  In-situ data augmentation will be incorporated with compute-intensive simulations, leading to discovery in multi- and inter-disciplinary, systems-wide, real-time recommendation systems. From instrumentation calibration, experimental and observable results, theoretical simulations, to validation and analysis---exascale computing will be brought to bear on health, transportation, environment, and social sciences. For example, climate and weather recommendation systems will be able to integrate models in storms prediction, rainfall, vegetation, farm growth, pollution, power usage, traffic flow, etc. \cite{storm,precipatation,yield,pollution,power}. 
As inputs and outputs from many sensor devices become ubiquitous, the importance of a scalable AI framework and an extensible workflow tool will grow. 

Challenges will rise from algorithmic approaches as HPC systems continue to expand and evolve. Multiple-precision implementation will be unavoidable. Setting the basis and preparing for product diversity from different vendors will be an important consideration of an AI framework, as well.
One challenge in performance is to reduce the
impact of communication while maintaining good convergence of the algorithm, such as SGD, in the DNN framework. In the Horovod
framework~\cite{horovod}, for example, the solution is to distribute the
training of neural network on large machines using an efficient gradient
reduction strategy referred to as ring-allreduce. Notably, in the 2018 Gordon-Bell award winning
paper~\cite{exascale-dnn-climate:18}, Horovod was used in the context
of TensorFlow to train two DNN models on large systems, including the
Summit machine, to detect extreme weather patterns from climate
data. Similarly, Horovod was used in~\cite{laanait2019exascale} to
train a large DNN model for tackling inverse
problems in imaging. Algorithmic improvements towards individual or combined synchronous, asynchronous, and pipeline approaches are essential to improve resource usage as well as convergence.

There is a trend in increasingly larg model sizes (especially in NLP---the latest model \cite{t-nlg} has 17 Billion parameters), and a need in scientific applications (e.g., geospatial imaging) to process larger-dimension inputs. Although there exist exascale deep learning applications \cite{exascale-dnn-climate:18, stemdl} on pre-exascale system such as Summit thanks to the Tensor Core technology, those use cases push the limit on large-batch training for data parallelism and are not generally applicable to exascale learning challenges. Early efforts \cite{mesh-tf, gpipe} on model parallelism have made progress, but are not yet mature or generic. An AI framework that can efficiently exploit various level of parallelisms (data parallel, model parallel, hyperparameter search, etc) will be in demand.     

\section{Conclusion}
Exascale systems are an important investment in the US. With exascale, we envision a stronger economy and improved quality of life. It will also lead to important scientific discovery and resolution of complex issues related to national security. Development in AI software and tools that scale well on these systems is important, and even more critical for AI frameworks that also work well across the existing spectrum of exascale applications. Although many challenges exist, a primary roadblock is the lack of direct collaborative effort, and a software platform that values performance as the foremost priority. In this paper, we present a unique set of AI tools and algorithms, as well as efforts between collegiate and ORNL researchers, demonstrating that there is a pathway to integrate and deploy machine learning to those ends---with emphasis on two major applications on exascale systems. 

\section{Acknowledgments}
This material is based upon work supported in part by the Laboratory Directed Research and Development program at the Oak Ridge National Laboratory, which is operated by UT-Battelle, LLC., for the U.S. Department of Energy under Contract DE-AC05-00OR22725. The work was partly supported by the Scientific Discovery through Advanced Computing (SciDAC) program funded by U.S. Department of Energy, Office of Science, Advanced Scientific Computing Research, with specific thanks to the FASTMath Institutes.

This work was conducted at the Joint Institute for Computational Sciences (JICS) and the Innovative Computing Laboratory (ICL), sponsored by the National Science Foundation (NSF), through NSF REU Award \#1659502 and NSF Award \#1709069. This work used hardware donations from NVIDIA as well as the Extreme Science and Engineering Discovery Environment (XSEDE), which is supported by NSF grant number ACI-1548562. Computational Resources are available through a XSEDE education allocation award TG-ASC170031.

\bibliographystyle{ieeetr}
\bibliography{base}

\end{document}